\documentclass[runningheads]{llncs}
\usepackage{eccv}

\usepackage{eccvabbrv}

\usepackage{graphicx}
\usepackage{booktabs}
\usepackage{multirow}
\usepackage[table]{xcolor}

\usepackage[accsupp]{axessibility}  % Improves PDF readability for those with disabilities.

\usepackage{hyperref}

\usepackage{orcidlink}

\begin{document}

\title{Following Motion for Sequential Modeling in Video Frame Interpolation}


\author{Jaehyun Park\inst{1}\orcidlink{0009-0008-0653-4077} \and
Nam Ik Cho\inst{1,2,3}\orcidlink{0000-0001-5297-4649}}

\authorrunning{J.~Park and N.~I.~Cho}

\institute{IPAI, Seoul National University \and
  Department of Electrical and Computer Engineering, INMC, Seoul National University \and
  SNUAILab\\
\email{\{jaep0805,nicho\}@snu.ac.kr}}

\maketitle

\begin{abstract}
State Space Models (SSMs) have surfaced as a promising architecture in Video Frame Interpolation (VFI), as they can capture long-range dependencies with linear computational complexity. However, their predefined scanning order limits their effectiveness in modeling the dynamic motion trajectories inherent in VFI problems. To tackle this challenge, we propose \textbf{M}otion-\textbf{G}uided \textbf{M}amba for \textbf{V}ideo \textbf{F}rame \textbf{I}nterpolation (MGMVFI), an adaptation of the selective state space model tailored explicitly for VFI. 
MGMVFI introduces Motion-Guided Serialization (MGS), which leverages optical flow to define a motion-adaptive 1D input order for the SSM. This aligns the causal state updates with semantically related tokens, enabling motion-consistent feature propagation, particularly for large and dynamic motions. 
Additionally, to mitigate the unreliable feature representations caused by inaccurate optical flow estimates, we introduce contextual synthesis that utilizes the surrounding spatial context for robust inter-frame feature synthesis.
These components are seamlessly integrated within our tailored Mamba architecture, which also employs a lightweight refinement block to enhance local detail reconstruction at a reduced computational cost.
Extensive experiments on standard VFI benchmarks demonstrate that MGMVFI achieves state-of-the-art performance, particularly on complex and dynamic motions, thereby establishing a new direction for sequence modeling in video interpolation.
\end{abstract}
\section{Introduction}
\label{sec:intro}

\begin{figure}[t]
  \centering
  \begin{tabular}[h]{ccc}
    \includegraphics[width=\textwidth]{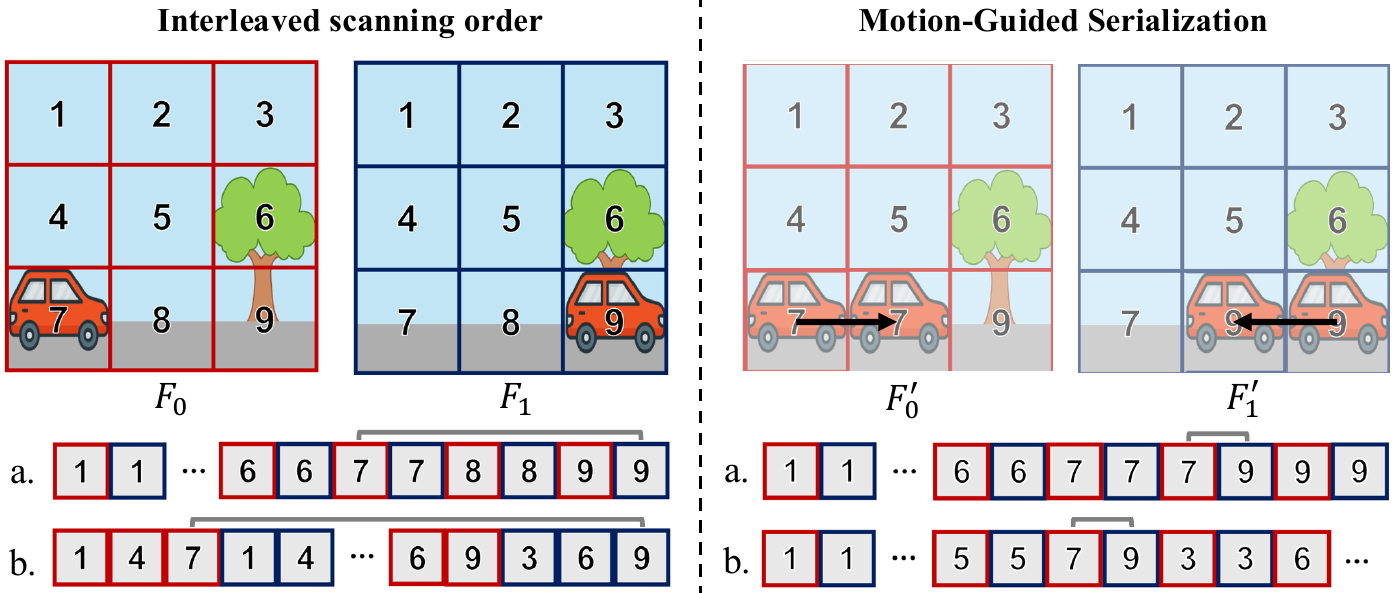}  \\

  \end{tabular}
  \caption{We present Motion-Guided Serialization (MGS), which uses optical flow to sample from motion-aligned feature maps $F'_0$ \& $F'_1$. MGS results in \textit{consecutive} semantically consistent tokens (Patch \textcolor{red}{7} \& \textcolor{blue}{9}) for \textbf{a.} Horizontal and \textbf{b.} Vertical scans. }
  \label{fig:scan_comparison}
\end{figure}

Video Frame Interpolation (VFI) is a long-standing task in computer vision that aims to synthesize intermediate frames between consecutive input frames in a video sequence.
This task plays a crucial role in numerous applications, such as video frame rate up-conversion \cite{bao2019depth, cheng2020video, choi2020channel, huang2022real}, slow-motion video generation \cite{bao2019depth, jiang2018super}, and novel view synthesis \cite{zhou2016view, kalantari2016learning, flynn2016deepstereo}.
By increasing temporal resolution, VFI improves visual smoothness and performance of downstream video tasks.

The effectiveness of deep learning-based VFI methods depends on how well the model captures the spatial and temporal dependencies in the input frame pair.
Conventional approaches have primarily relied on convolutional neural networks (CNNs) \cite{choi2020channel,lee2020adacof,huang2022real,bao2019depth} and Transformer-based architectures \cite{zhang2023extracting,shi2022video, lu2022video}.
However, CNNs suffer from limited receptive fields, making them less suitable for modeling large motions that span an extensive range.
Transformers, on the other hand, offer global attention mechanisms but suffer from quadratic computational complexity, limiting their scalability for high-resolution, real-time video processing.

Recently, state space models (SSMs), notably the Selective State Space (S6/ Mamba) family, have emerged as promising alternatives that combine the efficiency of convolutional operations with the dynamic modeling capability of recurrent structures \cite{gu2021combining, gu2021efficiently, gu2024mamba}.
S6-based architectures overcome the drawbacks of CNNs and Transformers by updating hidden states in a \textit{selective} and \textit{recurrent} manner, enabling the capture of long-range dependencies with linear complexity \cite{gu2024mamba}.
The inherent sequential nature of SSMs makes them highly dependent on the scanning order of the input features \cite{ibrahim2025survey}.
This sequence order is very pivotal, as it fundamentally dictates the flow of information through state updates, shaping the model's understanding of spatial and temporal relationships.

To process video inputs, prior works \cite{gu2024mamba, liu2024vmamba, li2024videomamba, guo2024mambair, guo2025mambairv2} have expanded 2D scanning strategies, such as horizontal, vertical, spiral scans, and Z-scans, by scanning each spatial coordinate sequentially for all images.
Furthermore, 3D scans, such as the interleaved \cite{zhang2024vfimamba} and Hilbert curve-based scans \cite{jeong2025lc}, have also been specifically developed to learn features that cover both spatial and temporal dimensions.
However, these pre-defined, motion-agnostic scanning paths are fundamentally misaligned with the dynamic nature of video.
As shown in Figure~\ref{fig:scan_comparison}, interleaved scanning \cite{zhang2024vfimamba} interleaves the columns of $F_0$ and $F_1$ to incorporate the temporal dimension. However, this simple approach often aggregates semantically inconsistent features across time. For instance, Patches \textcolor{red}{7} and \textcolor{blue}{9} represent features belonging to the same moving object. However, they have a large sequential distance (\textcolor{gray}{Gray} line) in the interleaved scan's input sequence, forcing the SSM to both estimate the motion trajectories and model the inter-frame features.
Moreover, the sequential distance depends on not only the magnitude of the motion but also the direction of the scan path, introducing additional variability that further hampers the estimation of motion trajectories.

Crucially, we reinterpret motion alignment as a \emph{serialization} problem for sequence models. Classic flow-based VFI methods \cite{jiang2018super, bao2019depth, niklaus2020softmax} estimate intermediate flows and visibility to warp features for spatial synthesis. In contrast, our goal is to define the \emph{1D input sequence order} for the SSM: motion trajectories determine how tokens are serialized, so the causal state updates occur between semantically related patches.
To this end, we propose a Motion-Guided Serialization (MGS) method that utilizes optical flow to dynamically rearrange input features along estimated motion paths.
We combine this with patch-level interleaved scans to ensure that adjacent patches in the SSM input correspond to semantically-related and temporally-coherent content.
The comparison is illustrated in Figure~\ref{fig:scan_comparison}.
Compared to interleaved scanning\cite{zhang2024vfimamba}, our motion-aligned serialization method produces a semantically consistent input sequence, demonstrated by the adjacent features of the moving car in the serialized SSM input sequence (Patch \textcolor{red}{7} and \textcolor{blue}{9}). Additionally, because we interleave at patch-level, the semantically-related features stay adjacent regardless of scan direction.

However, MGS's dependence on optical flow estimation inherently creates inaccuracies that arise from occlusion/disocclusion, abrupt motion and lighting changes. We therefore further design a complementary masked adapter module to handle areas with inaccurate flow with contextual synthesis.
Contextual synthesis identifies these information gaps and synthesizes robust features as replacements by leveraging the surrounding valid spatial context.
The resulting motion-aligned sequence provides a rich, coherent input that vastly relaxes the motion estimation and inter-frame modeling objective for the SSM. This allows the downstream SSM to efficiently handle temporal propagation followed by the modified Efficient Discriminative Frequency-domain FFN (EDFFN) for enhanced local detail reconstruction at reduced computational cost\cite{kong2025efficient}.

Our contribution can be summarized in threefold:
\begin{enumerate}
  \item \textbf{Motion-Guided Serialization (MGS)}: A novel serialization mechanism that leverages optical flow to dynamically define the SSM input sequence order along estimated motion paths, ensuring semantically consistent sequence modeling for complex and large-magnitude motions.
  \item \textbf{Contextual Synthesis}: A robust strategy to restore corrupted feature representations in regions with inaccurate or missing flow by leveraging valid surrounding spatial context to synthesize replacement features.
  \item \textbf{Adaptation of the Mamba architecture}: The integration of a modified Efficient Discriminative Frequency-domain FFN (EDFFN) into the Mamba framework to enhance local texture and detail reconstruction at a low cost.
\end{enumerate}

\section{Related Work}
\label{sec:relatedworks}
\textbf{Video Frame Interpolation.}
Early deep learning-based VFI methods were predominantly built on CNNs. These approaches can be broadly categorized into flow-based and kernel-based methods.
Flow-based VFI methods first estimate the optical flow between input frames and then warp these frames to the intermediate time step. Pioneering works, such as \cite{long2016learning}, introduced this concept, which has been followed by numerous refinements. These include direct estimation of intermediate flows \cite{liu2017video, jiang2018super}, bilateral motion estimation \cite{park2020bmbc, park2021asymmetric}, and the use of pre-trained contextual features to enhance the warping process \cite{niklaus2018context}. More recent efforts have focused on efficiency, such as combining estimation and refinement into a single encoder \cite{kong2022ifrnet} or using distillation \cite{huang2022real}.
Even in the modern landscape, advanced CNN architectures, such as SGM-VFI \cite{liu2024sparse}, continue to demonstrate competitive performance.
Kernel-based VFI methods, pioneered by \cite{niklaus2017video}, bypass explicit flow estimation.
Instead, they estimate spatially-adaptive convolution kernels that sample and weigh neighboring pixels from the input frames to directly synthesize the intermediate frame.
To overcome the limitations of fixed local kernels, subsequent works explored adaptive filters \cite{lee2020adacof} and deformable convolutions \cite{cheng2020video, cheng2021multiple} to better handle large motions and occlusions.

To address the limited receptive fields of CNNs and better model long-range dependencies, researchers began adopting the Transformer architecture. Initial works \cite{lu2022video, zhang2023extracting} demonstrated the power of attention mechanisms for VFI. The Transformer was also applied to kernel-based methods \cite{shi2022video} to address content-agnostic weights. Recently, EMA-VFI \cite{zhang2023extracting} has refined the Transformer architecture, achieving a balance of performance and efficiency that makes it a key state-of-the-art baseline.

A distinct line of research has focused on enhancing the perceptual quality of interpolated frames, particularly in challenging scenarios involving complex motion. Generative models, particularly Denoising Diffusion Probabilistic Models \cite{ho2020denoising}, have demonstrated remarkable potential. Works such as \cite{jain2024video} and \cite{danier2024ldmvfi} have demonstrated that generative priors can yield more realistic results, albeit at a significant computational cost. More recently, EDEN \cite{zhang2025eden} integrated a Transformer-based tokenizer with a DiT-style diffusion backbone, while Lyu et al. \cite{lyu2025tlb, lyu2024frame} adapted Brownian Bridge Diffusion Models \cite{li2023bbdm} to set a high standard for perceptual quality and robustness. In this work, we set generative models aside and focus on deterministic models and their reconstruction performance.

\noindent\textbf{State Space Models.}
State Space Models (SSMs) \cite{gu2021combining} were initially proposed for sequential signal modeling and later adapted for use in deep learning, utilizing HiPPO-based initialization and efficient structured representations. S4 \cite{gu2021efficiently} introduced parallelized updates to the hidden state, and S6 (Mamba) \cite{gu2024mamba} further advanced the design with selective gating and hardware-aware implementations.

In vision applications, SSMs have been adapted to 2D and video domains \cite{ibrahim2025survey}. Liu et al. \cite{liu2024vmamba} extended selective scanning to image patches, offering global receptive fields, while VideoMamba \cite{li2024videomamba} incorporated hierarchical bidirectional state updates for efficient video sequence modeling.
\cite{zhang2024motion, zhou2026mvssm} modeled motion-aware dynamic trajectories and temporal dependencies across frames to capture complex temporal dynamics in video sequences.
In image restoration, MambaIR \cite{guo2024mambair} introduced local enhancement and channel attention to address pixel forgetting, followed by semantic-based reordering for local-global reasoning \cite{guo2025mambairv2}.

For frame interpolation, VFIMamba \cite{zhang2024vfimamba} was the first to adopt S6-based modeling, employing interleaved frame scanning and curriculum-based training to outperform both CNN- and transformer-based baselines.
Afterward, LC-Mamba \cite{jeong2025lc} adopted the Hilbert-curve scan method with shifted windows to maintain spatial continuity across window boundaries and extended this to a 3D voxel-level scan for better spatiotemporal modeling between frames.
Our work builds on this line by introducing Motion-Guided Serialization. Unlike prior SSM-based models, which used fixed or heuristic scans, we leverage optical flow to dynamically rearrange scan paths along true motion trajectories, making the sequence modeling itself motion-adaptive.

\section{Proposed Method}
\label{sec:method}
\begin{figure*}[t]
  \centering
  \includegraphics[width=\textwidth]{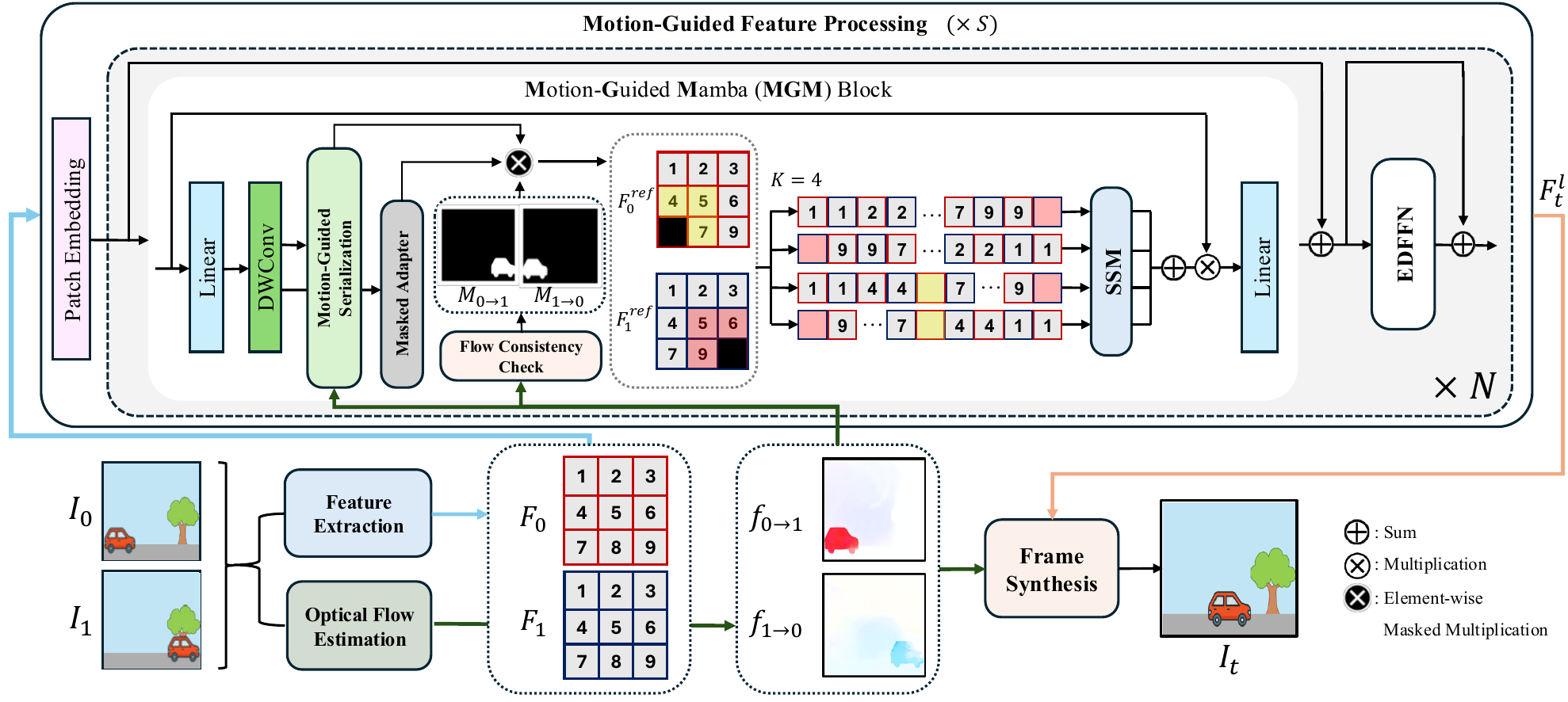}
  \caption{Overall pipeline of our model. Our MGMVFI block first rearranges input features based on their motion trajectory. It then augments features in regions with unreliable optical flow using spatial context from the surrounding (highlighted) patches. The Contextual Synthesis is visualized for a 3$\times$3 hollow kernel case.}
  \label{fig:overall_pipeline}
\end{figure*}

\subsection{Overview}
Our goal is to synthesize an intermediate frame $I_t \in \mathbb{R}^{H\times W \times 3}$ at an arbitrary timestep $t \in (0, 1)$ given two consecutive input frames, $I_0, I_1 \in \mathbb{R}^{H\times W \times 3}$. As illustrated in Figure~\ref{fig:overall_pipeline}, our model consists of three stages:
\begin{enumerate}
  \item \textbf{Feature Extraction and Optical Flow Estimation:} A shared CNN encoder first extracts multi-scale deep feature maps, $F_0$ and $F_1$, from the input frames $I_0$ and $I_1$. Furthermore, a pre-trained optical flow model is used to estimate the corresponding flow maps: $f_{0 \to 1}$ and $f_{1 \to 0}$.

  \item \textbf{Motion-Guided Feature Processing:}
    Our motion-guided pipeline processes the extracted features and optical flow with our proposed Motion-Guided Mamba for Video Frame Interpolation (MGMVFI) block.
    Within MGMVFI, the features are first processed using Motion-Guided Serialization and Contextual Synthesis to define a motion-adaptive serialization of the inputs.
    Then, the refined features are passed to our inter-frame feature modeling stage, which uses an SSM and the modified EDFFN block \cite{kong2025efficient} to efficiently capture local details and high-frequency information.

  \item \textbf{Frame Synthesis:} Finally, a frame synthesis module takes the motion-guided feature map $F_t$ and additional flow priors to synthesize the final RGB output frame, $I_t$. We utilize the established generation modules presented in \cite{zhang2023extracting} and \cite{huang2022real} to focus on the adaptation of SSMs.
\end{enumerate}

\subsection{Motion-Guided Serialization}

\label{sec:mgs}

The core of our MGMVFI block is \textbf{Motion-Guided Serialization (MGS)}, which reformulates how features are prepared for the SSM sequence model. Rather than using motion alignment for direct synthesis, MGS uses motion trajectories to define the serialization of features, i.e., the 1D token order fed into the SSM block. Concretely, we warp features to the intermediate grid so that standard raster scans over the resulting map correspond to motion-aligned sequences. This allows the causal flow of information to follow semantic motion, so SSM updates occur between related tokens and can focus on local residuals rather than global trajectory tracking. In practice, we serialize features by sampling along continuous motion trajectories, avoiding the limitations of fixed raster scans or linear blending that frequently produce ghosting artifacts near object boundaries.

To establish these trajectories, we must first estimate the intermediate flow fields originating from the target frame $I_t$. Given the bi-directional optical flow between the source frames, $f_{0 \to 1}$ and $f_{1 \to 0}$, we assume linear motion trajectories to scale the flow vectors. We employ forward splatting \cite{niklaus2020softmax}, denoted by $\mathcal{S}$, to push these scaled vectors onto the intermediate grid $p_t$:
\begin{equation}
  f_{t \to 0} = \mathcal{S}(-t \cdot f_{0 \to 1}), \qquad f_{t \to 1} = \mathcal{S}(-(1-t) \cdot f_{1 \to 0}),
\end{equation}
Then, calculate the source coordinates $p_0$ and $p_1$ by applying these intermediate flow fields to the target grid $p_t$:
\begin{equation}
  p_0 = p_t + f_{t \to 0}, \qquad p_1 = p_t + f_{t \to 1}
\end{equation}

Using these coordinates, we perform backward warping via bilinear sampling, denoted by $\mathcal{B}(\cdot,\cdot)$, to fetch the motion-aligned features from $F_0$ and $F_1$:

\begin{equation}
  F'_0 = \mathcal{B}(F_0, p_0), \qquad F'_1 = \mathcal{B}(F_1, p_1)
\end{equation}

Examples of the sampled $F'_0$ and $F'_1$ are illustrated in Figure~\ref{fig:scan_comparison}. By sampling from these motion-aligned feature maps, semantically consistent patches (e.g., Patch \textcolor{red}{7} and \textcolor{blue}{9}) are always consecutively ordered in the 1D sequence, simplifying the learning objective of the SSM to focus solely on inter-frame feature modeling.
To quantify this, we further define \textbf{Sequential Feature Similarity} on the 1D serialized sequence as
\begin{equation}
  \label{eq4}
  S = \frac{1}{N-1} \sum_{n=1}^{N-1} \exp\left( - \frac{\| \phi(z_{n+1}) - \phi(z_n) \|_2^2}{2\sigma^2} \right)
\end{equation}
where $N$ is the sequence length and $\phi(z_n)$ represents the feature representation of the \textit{n}-th token in the sequence.
We use this metric in our experiments (Table~\ref{tab:scanmethod}) to verify that MGS yields higher $S$, indicating smoother feature similarities between consecutive tokens and a simpler modeling target for the SSM.

\subsection{Contextual Synthesis for Unreliable Regions}

\label{sec:contextualsynthesis}

While backward warping populates the entire feature map, it fetches invalid or unreliable features in regions where accurate optical flow is unavailable. These failures occur not only due to physical occlusions and disocclusions, but also estimation inaccuracies caused by large motions or abrupt brightness changes. Consequently, the SSM receives invalid or corrupted tokens from these areas, resulting in substantial reconstruction errors. To address this, we introduce a novel \textbf{Contextual Synthesis} strategy designed to reconstruct these problematic regions by leveraging valid surrounding spatial information.

\noindent\textbf{Flow Consistency Check for Reliability Masking.}
To identify such regions, we first identify unreliable motion paths using a standard forward-backward consistency check on the optical flow maps $f_{0 \to 1}$ and $f_{1 \to 0}$. A motion path is considered unreliable if a point $p$ warped from $I_0$ to $I_1$ and back does not return to its original location:

\begin{equation}
  % \epsilon_{0 \to 1}(p) = || f_{0 \to 1}(p) - \mathcal{W}(f_{1 \to 0})||^2
  \epsilon_{0 \to 1}(p) = \Big|\Big| f_{0 \to 1}(p) + f_{1 \to 0}\big(p + f_{0 \to 1}(p)\big) \Big|\Big|_2^2
\end{equation}

A binary consistency mask, which we denote as $M_{0 \to 1} \in \{0,1\}^{H'\times W' \times 1}$, is produced by thresholding this squared L2-norm against the average flow magnitude. This identifies pixels where the motion path has failed:

\begin{equation}
  M_{0 \to 1}(p) =
  \begin{cases} 1 & \text{if } \epsilon_{0 \to 1}(p) > \alpha \cdot \overline{||f_{0 \to 1}||} \\ 0 & \text{otherwise}
  \end{cases}
\end{equation}

This check is particularly crucial for handling algorithmic failures, such as those caused by abrupt changes in brightness. When lighting variations violate the brightness constancy assumption of the flow network, it often defaults to a zero-magnitude vector. If applied to a moving object, backward warping with this zero vector fetches incorrect spatial locations (e.g., the static background instead of the moving object), causing severe ghosting artifacts. Because our consistency check evaluates the bidirectional agreement of the flow, it successfully flags these zero-vector failures, ensuring unreliable regions are masked regardless of whether they stem from physical occlusions or lighting variations. This bi-directional check produces two distinct binary masks:

\begin{itemize}
  \item $M_{0 \to 1}$: A forward consistency mask identifying regions in $I_0$ whose features become unreliable when mapped to $I_1$ (e.g. occluded regions).
  \item $M_{1 \to 0}$: A backward consistency mask identifying regions in $I_1$ whose features are unreliable when mapped back to $I_0$ (e.g. disoccluded regions).
\end{itemize}
These masks remain in the coordinates of their respective source frames.

\noindent\textbf{Contextual Synthesis via Masked Adapter.}
Synthesizing these unreliable regions is challenging because the corresponding features fetched from the opposite frame are corrupted. An intuitive example of this failure occurs during physical occlusion, as seen with Patch 9 in the interleaved scan of Figure~\ref{fig:scan_comparison}. Because the tree trunk visible in $F_0$ (Patch \textcolor{red}{9}) is covered by the moving car, features fetched from the same estimated location in $F_1$ (Patch \textcolor{blue}{9}) are incorrect. A straightforward solution is to use the mask $M_{0 \to 1}$ to simply zero out these incorrect features from $F'_1$. However, this forces the model to rely solely on the features from $F_0$, leading to low-quality synthesis characterized by blur and fading artifacts.
To fill this information gap, we introduce a masked adapter module $\mathcal{A}_1$ that replaces the corrupted features of $F'_1$ with valid surrounding context. As shown by the highlighted patches in Figure~\ref{fig:overall_pipeline}, $\mathcal{A}_1$ uses a $7 \times 7$ hollow convolution, which convolves over the surrounding patches while explicitly excluding the unreliable center patch:

\begin{equation}
  \label{eq7}
  F^{c}_1 = \mathcal{A}_1(F'_1) =  {Conv}_{1 \times 1}(\text{GELU}(ConvHollow_{7\times7}(F'_1))) + F'_1
\end{equation}

By excluding the identified failure patch, we guide the adapter $\mathcal{A}_1$ to leverage only the robust, surrounding valid context to supplement the missing feature.
% This supplemented representation is then utilized alongside the respective feature from $F_1$ by the SSM to interpolate the final feature.
A symmetric process is used to handle backward flow failures using a separate adapter $\mathcal{A}_0$ with the surrounding contextual features of $F'_0$.

\noindent\textbf{Refined Feature Composition.}
The final feature maps for each stream, $F^{ref}$, are composed by selecting either the motion-sampled feature $F'$ or the adapter's synthesized output $F^{c}$, guided by the appropriate mask:
\begin{equation}
  \label{eq8}
  F_{1}^{ref} = (1 - M_{0 \to 1}) \cdot F'_1 + M_{0 \to 1} \cdot F^{c}_1, \qquad
  F_{0}^{ref} = (1 - M_{1 \to 0}) \cdot F'_0 + M_{1 \to 0} \cdot F^{c}_0
\end{equation}
Finally, these two refined feature maps from \eqref{eq8} are fused with the original features via an additive connection before being passed to the main SSM block:
\begin{equation}
  \label{eq9}
  F_{1}^{fused} = F_1 + F_{1}^{ref}, \quad
  F_{0}^{fused} = F_0 + F_{0}^{ref}
\end{equation}

For each spatial location $i$ in the target grid, we interleave the corresponding features from $F_{0}^{fused}$ and $F_{1}^{fused}$ to form a sequence token input $z_i$:

\begin{equation}
  \label{eq10}
  z_i = [F_{0}^{fused}(i) \parallel F_{1}^{fused}(i)]
\end{equation}

where $\parallel$ denotes concatenation along the feature dimension. The final 1D sequence $Z = \{z_1, z_2, \dots, z_N\}$ is then fed into the Mamba SSM block.

\subsection{Inter-frame Feature Modeling}
\label{sec:featuremodeling}

The feature modeling in our pipeline utilizes the following two key components.

\noindent\textbf{SS2D Block.}
We adopt the SS2D block \cite{liu2024vmamba} as our SSM. Following its design, we perform $K=4$ different raster scans (horizontal, vertical, and their backward counterparts) on the motion-aligned input sequence $Z$.
Outputs are then aggregated and multiplied by the original input features to modulate intensity according to the spatiotemporal context captured by the SSM.
This gated representation then undergoes a final linear layer to produce the refined feature map. This adaptation allows robust estimation of motion trajectories and inter-frame feature modeling within a unified pipeline.

\noindent\textbf{EDFFN Block.}
While our MGS simplifies the SSM's objective by semantically aligning features, it also introduces computational overhead. Furthermore, a recent study \cite{ma2025tinyvim} suggests that the Mamba block tends to focus on low-frequency information, necessitating a complementary module to capture fine-grained details.
To offset the computational cost and simultaneously enhance local feature modeling, we modify and adapt the Efficient Discriminative Frequency-domain FFN (EDFFN). EDFFN is uniquely suited to our aligned features, as its frequency-domain gating mechanism selectively refines \textit{local} spatial details and enhances texture reconstruction, both of which are crucial for high-fidelity interpolation. On the other hand, the EDFFN's efficient design, which performs frequency operations on lower dimensions, provides significant computational savings.

\section{Implementation Details}
\noindent\textbf{Training Datasets and Details.}
We train our model on the Vimeo-90K \cite{xue2019video} triplet dataset, a large-scale collection known for its diverse motions. During training, the images are randomly cropped to $256 \times 256$ and augmented using horizontal, vertical, and temporal flipping, along with random rotations ($90^\circ, 180^\circ, 270^\circ$). We empirically set $S=2$ and $N=3$, repeating the feature processing for $\times8$ and $\times16$ downsampled features.

The model is trained for 150 epochs with a batch size of 8 on three NVIDIA RTX 3090 GPUs (approx. 3 days). We use the AdamW optimizer ($\beta_1 = 0.9$, $\beta_2 = 0.999$, weight decay $1 \times 10^{-4}$). The learning rate is initialized at $2 \times 10^{-4}$ and gradually decreased to $2 \times 10^{-5}$ using a cosine annealing schedule. To reduce computational overhead, we employ the pre-trained large RAFT \cite{teed2020raft} model and pre-compute the bi-directional flow maps. The flow consistency check threshold $\alpha$ is set to 1.1, which empirically yields the best results.

\noindent\textbf{Objective Functions.}
Our proposed model is trained by minimizing a composite objective function, $\mathcal{L}_{total}$, which combines a primary reconstruction loss and an auxiliary warping loss:
\begin{equation}
\mathcal{L}_{total} = \mathcal{L}_{lap} + \lambda_1 \mathcal{L}_{warp},
\end{equation}
where $\mathcal{L}_{lap}$ is the $\mathcal{L}_1$ loss computed between the Laplacian pyramids of the final interpolated frame $I_t$ and the ground-truth frame $I_t^{GT}$.
The warping loss, $\mathcal{L}_{warp}$, provides supervision on the image synthesized by the feature $F_t^s$ of each $s$-th scale of our multiscale architecture.
It is defined as the sum of Laplacian losses between the intermediate warped frames $I_t^s$ and the ground truth:
\begin{equation}
\mathcal{L}_{\text{warp}} = \sum_{s} \mathcal{L}_{\text{lap}}(I_{t}^{s}, I_{t}^{\text{GT}})
\end{equation}
Based on empirical validation, the weighting coefficient $\lambda_1$ is set to $0.5$.

\section{Experiment Results}
We compare our model against recent state-of-the-art (SotA) VFI models. As a deterministic model, we specifically choose to compare against CNN-based \cite{bao2019depth, lee2020adacof, choi2020channel, niklaus2020softmax, sim2021xvfi, hu2022many, huang2022real, kong2022ifrnet, li2023amt, liu2024sparse}, Transformer-based \cite{zhang2023extracting, zhang2025eden}, and S6-based \cite{zhang2024vfimamba, jeong2025lc} models. The metrics reported for LC-Mamba are from its balanced model (Ours-B).
For a more comprehensive qualitative assessment, we also include the diffusion-based generative model EDEN \cite{zhang2025eden}. As a generative model that focuses on perceptual metrics, EDEN serves as a strong reference for qualitative performance.

\noindent\textbf{Evaluation Datasets.} Our model is evaluated on four test benchmarks.
Vimeo90K \cite{xue2019video} (448$\times$256) and UCF101 \cite{soomro2012ucf101} (256$\times$256) consist of low-resolution real-world scenes, composed of varying motion magnitude. SNU-FILM \cite{choi2020channel} is a higher resolution dataset (1280$\times$720) that is divided into four difficulty splits(Easy, Medium, Hard, and Extreme) depending on the motion magnitude. Xiph\cite{montgomery1994xiph} consists of 4K-resolution frames, which we followed \cite{niklaus2020softmax} to reproduce 2K-resolution frames and evaluate on both.

\subsection{Quantitative Evaluation}
\begin{table*}[ht!]
\centering
\caption{Quantitative Evaluation across three benchmark datasets. The best and second best results are indicated by \textbf{bold} and \underline{underline}.}
\label{tab:quant_comparison_full}
\resizebox{\linewidth}{!}{%
\begin{tabular}{cccccccc}
\toprule
& Train & \multirow{2}{*}{Vimeo-90K \cite{xue2019video}} & \multirow{2}{*}{UCF101 \cite{soomro2012ucf101}} & \multicolumn{4}{c}{SNU-FILM \cite{choi2020channel}} \\
\cline{5-8} & Dataset &  & & Easy & Medium & Hard & Extreme \\ \midrule
DAIN \cite{bao2019depth} & V & 34.71/0.9756 & 34.99/0.9683 & 39.73/0.9902 & 35.46/0.9780 & 30.17/0.9335 & 25.09/0.8584 \\
AdaCof \cite{lee2020adacof} & V & 34.47/0.9730 & 34.90/0.9680 & 39.80/0.9900 & 35.05/0.9754 & 29.46/0.9244 & 24.31/0.8439 \\
CAIN \cite{choi2020channel} & V & 34.65/0.9730 & 34.91/0.9690 & 39.89/0.9900 & 35.61/0.9776 & 29.90/0.9292 & 24.78/0.8507 \\
Softsplat \cite{niklaus2020softmax}& V & 36.13/0.9805 & 35.39/0.9697 & 40.26/0.9911 & 36.07/0.9798 & 30.53/0.9365 & 25.16/0.8604 \\
XVFI \cite{sim2021xvfi}& V & 35.09/0.9759 & 35.17/0.9685 & 39.93/0.9907 & 35.37/0.9782 & 29.58/0.9276 & 24.17/0.8450 \\
M2M-VFI \cite{hu2022many}& V & 35.47/0.9778 & 35.28/0.9694 & 39.66/0.9904 & 35.74/0.9794 & 30.30/0.9360 & 25.08/0.8604 \\
RIFE \cite{huang2022real}& V & 35.61/0.9779 & 35.28/0.9690 & 39.80/0.9903 & 35.76/0.9787 & 30.36/0.9351 & 25.27/0.8601 \\
IFRNet-L \cite{kong2022ifrnet}& V & 36.20/0.9808 & 35.42/0.9698 & 40.10/0.9906 & 36.12/0.9797 & 30.63/0.9368 & 25.26/0.8609 \\
AMT-L \cite{li2023amt}& V & 36.35/0.9815 & 35.39/0.9698 & 39.95/\textbf{0.9913} & 36.09/\textbf{0.9805} & 30.75/0.9384 & 25.41/0.8638 \\
AMT-G \cite{li2023amt}& V & 36.53/0.9817 & 35.41/0.9699 & 39.88/\textbf{0.9913} & 36.12/\textbf{0.9805} & 30.78/0.9385 & 25.43/0.8644 \\
SGM-VFI \cite{liu2024sparse} & V & 35.81/0.9793 & 35.34/0.9693 & 40.14/0.9907 & 36.06/0.9795 & 30.81/0.9375 & 25.59/0.8646 \\
EMA-VFI-S \cite{zhang2023extracting} & V & 36.07/0.9797 & 35.34/0.9696 & 39.81/0.9906 & 35.88/0.9795 & 30.69/0.9375 & 25.47/0.8632 \\
EMA-VFI \cite{zhang2023extracting}& V & \underline{36.64}/\underline{0.9819} & \underline{35.48}/0.9701 & 39.98/0.9910 & 36.09/0.9801 & 30.94/0.9392 & 25.69/0.8661 \\
EDEN \cite{zhang2025eden} & LAVIB & 32.66/0.9587 & 34.92/0.9676 & 38.69/0.9879 & 34.66/0.9744 & 29.59/0.9279 & 24.94/0.8536 \\
VFIMamba \cite{zhang2024vfimamba}& V+X & \underline{36.64}/\underline{0.9819} & 35.45/\underline{0.9702} & \textbf{40.51}/\underline{0.9912} & \textbf{36.40}/\textbf{0.9805} & \underline{30.99}/\underline{0.9401} & \underline{25.79}/\underline{0.8682}  \\ 
LCMamba(B) \cite{jeong2025lc} & V & 36.43/0.9813 & 35.39/0.9698 & 40.07/0.9909 & 36.08/0.9801 & 30.59/0.9375 & 25.35/0.8630  \\ \midrule
Ours & V & \textbf{36.67}/\textbf{0.9820} & \textbf{35.53}/\textbf{0.9710} & \underline{40.45}/0.9910 & \underline{36.38}/\underline{0.9804} & \textbf{31.04}/\textbf{0.9411} & \textbf{25.86}/\textbf{0.8732} \\
\bottomrule
\end{tabular}
}
\end{table*}

To quantitatively evaluate our model, we report PSNR and SSIM, the standard reconstruction benchmarks in VFI literature. As shown in Table~\ref{tab:quant_comparison_full}, our model achieves state-of-the-art (SotA) performance on the Vimeo-90K and UCF101 datasets across various motion magnitudes, exceeding prior methods. The primary advantage of MGMVFI is most evident on the challenging SNU-FILM benchmark. While it performs competitively on Easy and Medium splits, MGMVFI establishes new SotA results on the Hard and Extreme splits, which feature significant, complex motion. On the Extreme split, MGMVFI outperforms the next-best S6-based model, VFIMamba, by 0.07 dB, confirming the robustness of our motion-aware approach for high-fidelity synthesis.

\begin{table*}[ht!]
\centering
    \captionof{table}{ Quantitative evaluation on Xiph \cite{montgomery1994xiph} dataset and efficiency comparison. Runtimes (RT) are measured for a 2048$\times$1024 resolution input on an RTX A6000.}
    \label{tab:xiph}
    \resizebox{0.6\textwidth}{!}{%
      \begin{tabular}{lcccccc}
      \toprule
       & \multicolumn{2}{c}{Xiph 2K} & \multicolumn{2}{c}{Xiph 4K} & Params(M) & RT(ms)\\ \midrule
      SGM-VFI & 36.57 & 0.9424 & 34.23 & 0.9021 & 20.8 & 1168.9\\
      EMA-VFI & 36.74 & 0.9445 & 34.54 & 0.9054 & 65.6 & 332.3\\
      EDEN & 33.36 & 0.9043 & 26.84 & 0.7303 & 157.8 & 1304.4\\
      VFIMamba & \underline{37.13} & 0.9451 & \underline{34.62} & \textbf{0.9059} & 66.1 & 414.7 \\
      LCMamba(B) & 36.90 & \textbf{0.9456} & 34.26 & 0.9040 & 16.2 & 785.6\\ \midrule
      \textbf{Ours} & \textbf{37.15} & \underline{0.9454} & \textbf{34.65} & \textbf{0.9059} & 78.4 & 427.5 \\
      \bottomrule
      \end{tabular}
    }
\end{table*}

The efficiency and high-resolution scalability of MGMVFI are further underscored by the Xiph benchmark results in Table~\ref{tab:xiph}. The performance on Xiph 4K highlights the importance of trajectory-aware serialization. While fixed scanning orders often fail to capture temporal dependencies at high resolutions where related features are spatially distant, MGMVFI dynamically aligns the sequence with the true motion path to effectively preserve structural integrity and fine textures. Furthermore, despite the motion-guided serialization step, our model maintains a competitive inference runtime of 427.5 ms. This efficiency stems from the EDFFN module, which selectively refines high-frequency details in the frequency domain, allowing for a streamlined sequence-modeling backbone.

\subsection{Qualitative Evaluation}
\begin{figure*}[t]
  \centering
  \includegraphics[width=\textwidth]{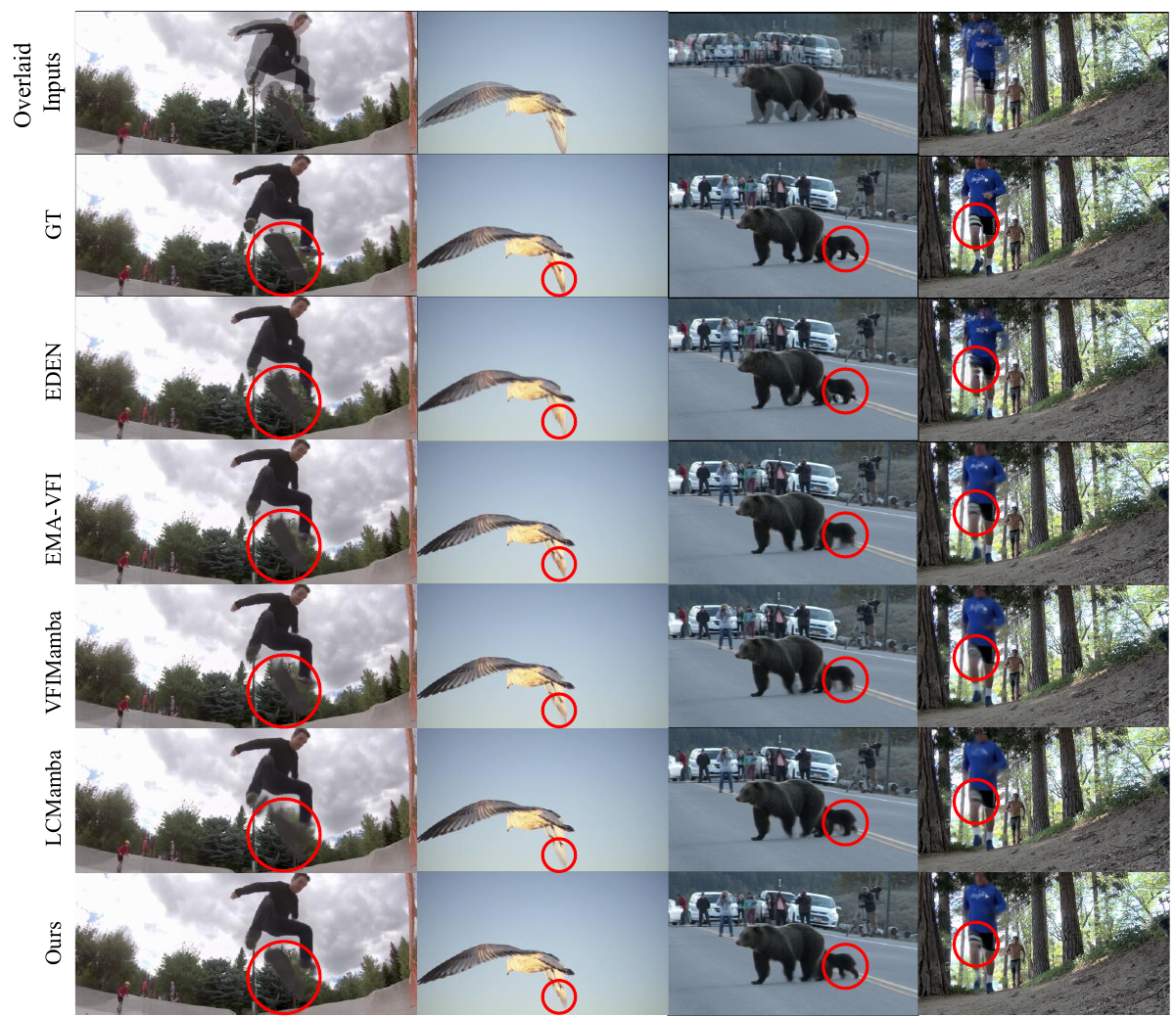}
  \caption{Visual comparison of the interpolated results from the SNU-FILM \cite{choi2020channel} Hard and Vimeo90K \cite{xue2019video} dataset. Best viewed zoomed in.}
  \label{fig:qual_comparison}
\end{figure*}

Figure~\ref{fig:qual_comparison} provides a qualitative comparison against leading VFI models using challenging scenes from the Vimeo90K and SNU-FILM Hard datasets. Scene 1 (columns 1–2) illustrates a rapid-motion example where the motion-agnostic scanning orders of prior S6-based models fail. Their fixed interleaved and Hilbert-curve scans provide semantically inconsistent patches, causing structural distortion. In contrast, our MGS module processes features along the motion path, providing a coherent sequence that simplifies the SSM's objective and produces structurally sound, sharp interpolations. Scene 2 (column 3) highlights a difficult disocclusion scenario. Baselines, including prior S6 models and EMA-VFI, exhibit prominent ghosting and background blur due to insufficient information in newly revealed regions. Conversely, our results demonstrate the effectiveness of the Contextual Synthesis mechanism; the masked adapter synthesizes robust features from the surrounding valid spatial context. Combined with our modified EDFFN, the model handles information gaps to produce a crisp, artifact-free background. Scene 3 (column 4) demonstrates our model's superiority in reconstructing high-frequency details. While baselines suffer from blur or washout artifacts that sacrifice faithfulness—as seen in the generative EDEN model—our model renders the sharpest textures. This performance is attributable to the EDFFN module, whose frequency-domain gating is uniquely suited for texture reconstruction, as validated by the 0.14 dB gain in Table~\ref{tab:modules}.

\subsection{Ablation Study}

\noindent\textbf{Progressive Module Analysis.}
\begin{figure*}[t]
  \centering
  \includegraphics[width=\textwidth]{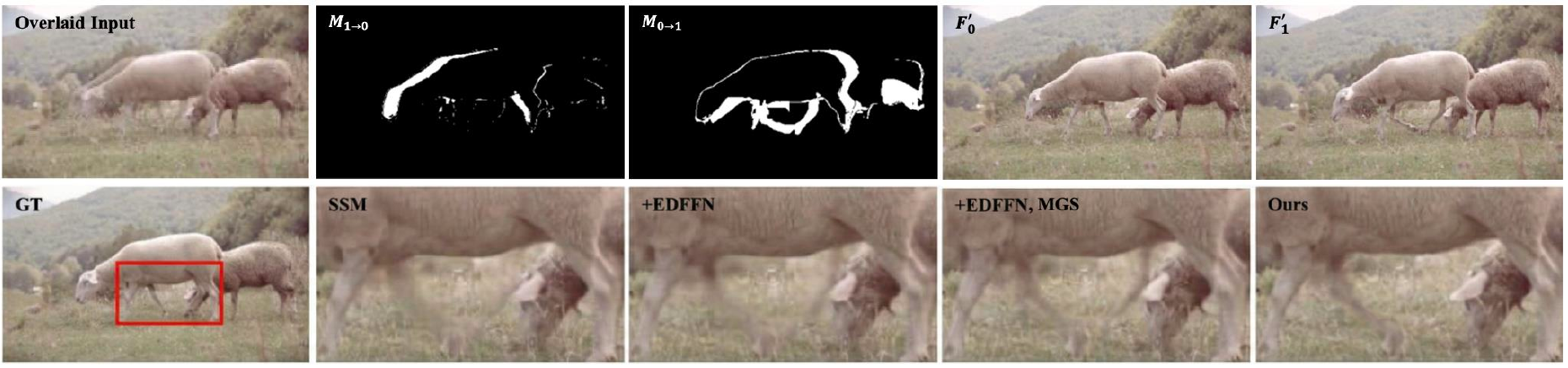}
  \caption{Visual comparison of the progressive module analysis from Vimeo-90K \cite{xue2019video}.}
  \label{fig:ab_prog}
\end{figure*}

We conduct a progressive ablation study on the Vimeo-90K, UCF101, and SNU-FILM Hard datasets to validate the contributions of our proposed modules, with quantitative results and visual comparisons provided in Table~\ref{tab:modules} and Figure~\ref{fig:ab_prog}, respectively. We begin with a baseline model consisting of core SSM blocks, which struggles to handle dynamic movement, resulting in significant blur and ghosting artifacts. Integrating the modified EDFFN block yields a consistent improvement in sharpness across all benchmarks, raising the PSNR on the SNU-FILM Hard split to 29.80 dB. The addition of our core MGS method provides a substantial performance leap, increasing the SNU-FILM Hard PSNR by 0.44 dB to 30.24 dB. Visually, the +EDFFN and MGS configuration produces much clearer images with reduced structural artifacts, demonstrating the effectiveness of aligning the serialization order with estimated motion paths. Finally, introducing the Contextual Synthesis module yields the most significant gain, boosting the SNU-FILM Hard PSNR to 31.04 dB and the Vimeo-90K PSNR to 36.67 dB. This full model effectively reconstructs complex occlusions, such as a sheep's head, while correctly eliminating disoccluded background artifacts (hind leg) as guided by the consistency mask $M_{1 \to 0}$.

\begin{table}[!htb]
    \centering
    \begin{minipage}{0.48\textwidth}
        \centering
        \caption{Quantitative report of the progressive module analysis}
        \label{tab:modules}
        \resizebox{\linewidth}{!}{
        \begin{tabular}{lcccccc}
        \toprule
          \multirow{2}{*}{Scan method}  & \multicolumn{2}{c}{Vimeo-90K\cite{xue2019video}} &  \multicolumn{2}{c}{UCF101\cite{soomro2012ucf101}} &  \multicolumn{2}{c}{Hard\cite{choi2020channel}}\\ \cmidrule{2-7}
          &PSNR  &SSIM & PSNR & SSIM & PSNR & SSIM \\ \midrule
        SSM &35.96&0.9793&35.37&0.9699&29.71&0.9293 \\
        + EDFFN &36.10&0.9797&35.38&0.9701&29.80&0.9300 \\
        + MGS &36.26&0.9802&35.56&0.9713&30.24&0.9345 \\
        + Cont. Synth.  &36.67&0.9820&35.53&0.9710&31.04&0.9411 \\
        \bottomrule
        \end{tabular}
        }
    \end{minipage}%
    \hfill
    \begin{minipage}{0.48\textwidth}
        \centering
        \caption{Performance comparison with different optical flow backbones on SNU-FILM Hard \cite{choi2020channel}}
        \label{tab:opticalflow}
        \resizebox{\linewidth}{!}{%
        \begin{tabular}{l|cc|cc}
        \toprule
          Flow Backbone  & RT(ms) & Params(M) & PSNR & SSIM \\ \midrule
          LiteFlowNet & 185.12 & 4.86 & 30.89 & 0.9405 \\
          PWC-Net & 218.23 & 9.37 & 30.70 & 0.9388 \\
          RAFT - Small & 277.88 & 0.99 & 30.92 & 0.9408 \\
          RAFT - Large & 415.90 & 5.26 & 31.04 & 0.9411 \\
        \bottomrule
        \end{tabular}
        }
    \end{minipage}
\end{table}

\begin{figure}[htbp]
  \centering
  
  \begin{minipage}[c]{0.53\textwidth}
    \centering
    \captionof{table}{Comparison of MGS against previous scanning methods}
    \label{tab:scanmethod}
    \resizebox{\linewidth}{!}{%
    \begin{tabular}{lccccc}
    \toprule
    \multirow{2}{*}{Scan method} & \multirow{2}{*}{$S \uparrow$} & \multicolumn{2}{c}{Vimeo-90K\cite{xue2019video}} & \multicolumn{2}{c}{Hard\cite{choi2020channel}} \\ \cmidrule{3-6}
    && PSNR & SSIM & PSNR & SSIM \\ \midrule
    Sequential & 0.15 & 35.39  & 0.9791  & 29.45  & 0.9336  \\
    Spiral     & 0.12 & 35.17  & 0.9781  & 28.83  & 0.9302  \\
    Z-scan     & 0.29 & 35.50  & 0.9792  & 29.79  & 0.9335  \\
    Interleaved & 0.35 & 36.56  & 0.9817  & 30.88  & 0.9390  \\
    Hilbert Curve & 0.33 & 36.45  & 0.9813  & 30.57  & 0.9376  \\ \midrule
    Ours (MGS) & 0.46 & 36.67  & 0.9820  & 31.04  & 0.9411  \\
    \bottomrule
\end{tabular}
    }
  \end{minipage}
  \hfill 
  \begin{minipage}[c]{0.44\textwidth}
      \centering
    \includegraphics[width=\textwidth]{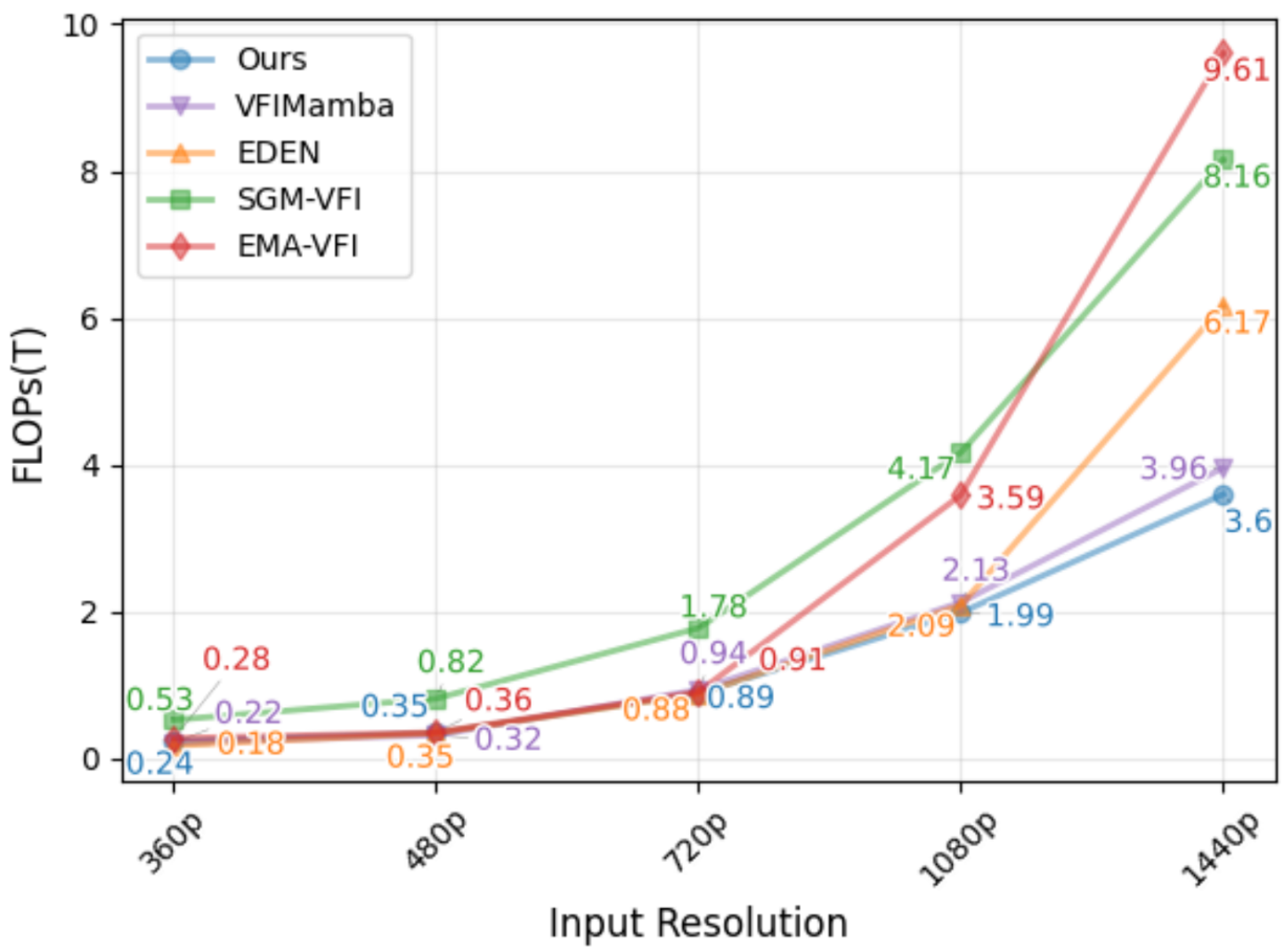}
      \captionof{figure}{Computation comparison on different input resolution}
    \label{fig:eff}
  \end{minipage}
\end{figure}

\noindent\textbf{Optical Flow Sensitivity Analysis.}
For our model, we employed the RAFT-Large model \cite{teed2020raft} to generate the flow maps $f_{0\rightarrow1}$ and $f_{1\rightarrow0}$. In this section, we examine changes in performance when using flow estimators \cite{teed2020raft, sun2018pwc, hui2018liteflownet} at varying levels of accuracy and computational cost.
As shown in Table~\ref{tab:opticalflow}, while using RAFT-Large yields the highest fidelity, our model maintains robust performance even with lightweight estimators such as RAFT-Small. The performance drop is marginal (approx. 0.12dB), indicating that MGMVFI is not reliant on perfect flow estimation.
This is further supported by the performance of our model across different scan methods presented below (Table~\ref{tab:scanmethod}), indicating that the SSM itself possesses inherent motion-modeling capabilities.
Furthermore, utilizing a smaller flow backbone significantly reduces the overall inference latency, offering a flexible trade-off for resource-constrained applications.

\noindent\textbf{Scan methods.}
Table~\ref{tab:scanmethod} evaluates our Motion-Guided Serialization (MGS) against conventional single-image (Sequential, Spiral, Z-scan) and video-specific (Interleaved, Hilbert Curve) scanning methods. To quantify the semantic continuity of the 1D serialized sequence, we introduce the Sequential Feature Similarity ($S$) metric (\ref{eq4}) measured across 100 randomly sampled Vimeo90K images. Our results show that MGS significantly outperforms all baselines, particularly in complex motion scenarios. Specifically, MGS achieves a superior $S$ of 0.46— a 31\% improvement over the next best method (Interleaved). This semantic consistency directly translates into reconstruction accuracy, with MGS achieving a SotA PSNR of 31.04 dB on the SNU-FILM Hard benchmark. This highlights MGS's semantic consistency, simplifying the sequence modeling task for the S6 block and enabling more accurate local feature reconstruction.

\noindent\textbf{Computation scalability.}
Finally, the scalability of the S6 architecture is evaluated by measuring computational load against input resolution in Figure ~\ref{fig:eff}. While competitive at lower scales, the MGMVFI block exhibits a clear advantage at high resolutions, where its linear scaling results in substantially lower resource consumption than other architectures.

\section{Limitations}

Despite MGMVFI's strong performance for SSM-based interpolation, several avenues for improvement remain.
MGMVFI relies on external optical flow for MGS; flow errors can corrupt token ordering and propagate through the SSM. While contextual synthesis is beneficial, it does not address the core problem.
Furthermore, MGS uses a single motion-adaptive serialization, which can underperform in scenes with multiple objects and small motions.
We leave new research directions, such as joint flow refinement and multi-trajectory serialization, for future work.

\section{Conclusion}
We have proposed MGMVFI — Motion-Guided Mamba for Video Frame Interpolation — a motion-aware framework that redefines how S6 models handle VFI by moving beyond fixed scanning patterns. By implementing the MGS module, we leveraged optical flow to create a motion-adaptive serialization that preserved temporal coherence and streamlined interpolation. This was complemented by contextual synthesis to mitigate occlusions and an optimized EDFFN block that improved computational efficiency while restoring fine details. MGMVFI achieved state-of-the-art performance on challenging benchmarks, proving the efficacy of motion-guided serialization. We anticipate that this motion-guided serialization approach will provide a scalable foundation for broader video synthesis and understanding tasks.

\section*{Acknowledgements}
This work was supported by Samsung Electronics Co., Ltd(IO251211-14315-01).

% \clearpage
\bibliographystyle{splncs04}
\bibliography{main}

@String(AAAI  = {AAAI})

@String(TOG   = {ACM Trans. Graph.})

@String(TOG   = {ACM TOG})

@inproceedings{zhou2026mvssm,
  title={MVSSM: Motion-aware Visual State Space Model for Efficient Video Deblurring},
  author={Zhou, Chen and Wu, Tao and Liu, Wei and Wu, Xi and Fu, Ying},
  booktitle={Proceedings of the IEEE/CVF Conference on Computer Vision and Pattern Recognition},
  pages={4855--4865},
  year={2026}
}

@inproceedings{zhang2024motion,
  title={Motion mamba: Efficient and long sequence motion generation},
  author={Zhang, Zeyu and Liu, Akide and Reid, Ian and Hartley, Richard and Zhuang, Bohan and Tang, Hao},
  booktitle={European Conference on Computer Vision},
  pages={265--282},
  year={2024},
  organization={Springer}
}

@article{zhang2024vfimamba,
  title={Vfimamba: Video frame interpolation with state space models},
  author={Zhang, Guozhen and Liu, Chuxnu and Cui, Yutao and Zhao, Xiaotong and Ma, Kai and Wang, Limin},
  journal={Advances in Neural Information Processing Systems},
  volume={37},
  pages={107225--107248},
  year={2024}
}

@inproceedings{jeong2025lc,
  title={LC-Mamba: Local and Continuous Mamba with Shifted Windows for Frame Interpolation},
  author={Jeong, Min Wu and Rhee, Chae Eun},
  booktitle={Proceedings of the Computer Vision and Pattern Recognition Conference},
  pages={17671--17681},
  year={2025}
}

@inproceedings{zhang2025eden,
  title={Eden: Enhanced diffusion for high-quality large-motion video frame interpolation},
  author={Zhang, Zihao and Chen, Haoran and Zhao, Haoyu and Lu, Guansong and Fu, Yanwei and Xu, Hang and Wu, Zuxuan},
  booktitle={Proceedings of the Computer Vision and Pattern Recognition Conference},
  pages={2105--2115},
  year={2025}
}

@inproceedings{zhang2023extracting,
  title={Extracting motion and appearance via inter-frame attention for efficient video frame interpolation},
  author={Zhang, Guozhen and Zhu, Yuhan and Wang, Haonan and Chen, Youxin and Wu, Gangshan and Wang, Limin},
  booktitle={Proceedings of the IEEE/CVF Conference on Computer Vision and Pattern Recognition},
  pages={5682--5692},
  year={2023}
}

@inproceedings{liu2024sparse,
  title={Sparse global matching for video frame interpolation with large motion},
  author={Liu, Chunxu and Zhang, Guozhen and Zhao, Rui and Wang, Limin},
  booktitle={Proceedings of the IEEE/CVF conference on computer vision and pattern recognition},
  pages={19125--19134},
  year={2024}
}

@article{soomro2012ucf101,
  title={Ucf101: A dataset of 101 human actions classes from videos in the wild},
  author={Soomro, Khurram and Zamir, Amir Roshan and Shah, Mubarak},
  journal={arXiv preprint arXiv:1212.0402},
  year={2012}
}

@article{xue2019video,
  title={Video enhancement with task-oriented flow},
  author={Xue, Tianfan and Chen, Baian and Wu, Jiajun and Wei, Donglai and Freeman, William T},
  journal={International Journal of Computer Vision},
  volume={127},
  number={8},
  pages={1106--1125},
  year={2019},
  publisher={Springer}
}

@inproceedings{choi2020channel,
  title={Channel attention is all you need for video frame interpolation},
  author={Choi, Myungsub and Kim, Heewon and Han, Bohyung and Xu, Ning and Lee, Kyoung Mu},
  booktitle={Proceedings of the AAAI conference on artificial intelligence},
  volume={34},
  pages={10663--10671},
  year={2020}
}

@article{liu2024vmamba,
  title={Vmamba: Visual state space model},
  author={Liu, Yue and Tian, Yunjie and Zhao, Yuzhong and Yu, Hongtian and Xie, Lingxi and Wang, Yaowei and Ye, Qixiang and Jiao, Jianbin and Liu, Yunfan},
  journal={Advances in neural information processing systems},
  volume={37},
  pages={103031--103063},
  year={2024}
}

@inproceedings{guo2024mambair,
  title={Mambair: A simple baseline for image restoration with state-space model},
  author={Guo, Hang and Li, Jinmin and Dai, Tao and Ouyang, Zhihao and Ren, Xudong and Xia, Shu-Tao},
  booktitle={European conference on computer vision},
  pages={222--241},
  year={2024},
  organization={Springer}
}

@inproceedings{guo2025mambairv2,
  title={Mambairv2: Attentive state space restoration},
  author={Guo, Hang and Guo, Yong and Zha, Yaohua and Zhang, Yulun and Li, Wenbo and Dai, Tao and Xia, Shu-Tao and Li, Yawei},
  booktitle={Proceedings of the Computer Vision and Pattern Recognition Conference},
  pages={28124--28133},
  year={2025}
}

@inproceedings{li2024videomamba,
  title={Videomamba: State space model for efficient video understanding},
  author={Li, Kunchang and Li, Xinhao and Wang, Yi and He, Yinan and Wang, Yali and Wang, Limin and Qiao, Yu},
  booktitle={European conference on computer vision},
  pages={237--255},
  year={2024},
  organization={Springer}
}

@inproceedings{gu2024mamba,
  title={Mamba: Linear-time sequence modeling with selective state spaces},
  author={Gu, Albert and Dao, Tri},
  booktitle={First conference on language modeling},
  year={2024}
}

@article{gu2021efficiently,
  title={Efficiently modeling long sequences with structured state spaces},
  author={Gu, Albert and Goel, Karan and R{\'e}, Christopher},
  journal={arXiv preprint arXiv:2111.00396},
  year={2021}
}

@inproceedings{danier2024ldmvfi,
  title={Ldmvfi: Video frame interpolation with latent diffusion models},
  author={Danier, Duolikun and Zhang, Fan and Bull, David},
  booktitle={Proceedings of the AAAI Conference on Artificial Intelligence},
  volume={38},
  pages={1472--1480},
  year={2024}
}

@inproceedings{jain2024video,
  title={Video interpolation with diffusion models},
  author={Jain, Siddhant and Watson, Daniel and Tabellion, Eric and Poole, Ben and Kontkanen, Janne and others},
  booktitle={Proceedings of the IEEE/CVF Conference on Computer Vision and Pattern Recognition},
  pages={7341--7351},
  year={2024}
}

@inproceedings{huang2022real,
  title={Real-time intermediate flow estimation for video frame interpolation},
  author={Huang, Zhewei and Zhang, Tianyuan and Heng, Wen and Shi, Boxin and Zhou, Shuchang},
  booktitle={European Conference on Computer Vision},
  pages={624--642},
  year={2022},
  organization={Springer}
}

@inproceedings{teed2020raft,
  title={Raft: Recurrent all-pairs field transforms for optical flow},
  author={Teed, Zachary and Deng, Jia},
  booktitle={European conference on computer vision},
  pages={402--419},
  year={2020},
  organization={Springer}
}

@inproceedings{kong2025efficient,
  title={Efficient visual state space model for image deblurring},
  author={Kong, Lingshun and Dong, Jiangxin and Tang, Jinhui and Yang, Ming-Hsuan and Pan, Jinshan},
  booktitle={Proceedings of the Computer Vision and Pattern Recognition Conference},
  pages={12710--12719},
  year={2025}
}

@article{gu2021combining,
  title={Combining recurrent, convolutional, and continuous-time models with linear state space layers},
  author={Gu, Albert and Johnson, Isys and Goel, Karan and Saab, Khaled and Dao, Tri and Rudra, Atri and R{\'e}, Christopher},
  journal={Advances in neural information processing systems},
  volume={34},
  pages={572--585},
  year={2021}
}

@inproceedings{cheng2020video,
  title={Video frame interpolation via deformable separable convolution},
  author={Cheng, Xianhang and Chen, Zhenzhong},
  booktitle={Proceedings of the AAAI Conference on Artificial Intelligence},
  volume={34},
  pages={10607--10614},
  year={2020}
}

@article{cheng2021multiple,
  title={Multiple video frame interpolation via enhanced deformable separable convolution},
  author={Cheng, Xianhang and Chen, Zhenzhong},
  journal={IEEE Transactions on Pattern Analysis and Machine Intelligence},
  volume={44},
  number={10},
  pages={7029--7045},
  year={2021},
  publisher={IEEE}
}

@inproceedings{lee2020adacof,
  title={Adacof: Adaptive collaboration of flows for video frame interpolation},
  author={Lee, Hyeongmin and Kim, Taeoh and Chung, Tae-young and Pak, Daehyun and Ban, Yuseok and Lee, Sangyoun},
  booktitle={Proceedings of the IEEE/CVF Conference on Computer Vision and Pattern Recognition},
  pages={5316--5325},
  year={2020}
}

@inproceedings{niklaus2017video,
  title={Video frame interpolation via adaptive separable convolution},
  author={Niklaus, Simon and Mai, Long and Liu, Feng},
  booktitle={Proceedings of the IEEE international conference on computer vision},
  pages={261--270},
  year={2017}
}

@inproceedings{bao2019depth,
  title={Depth-aware video frame interpolation},
  author={Bao, Wenbo and Lai, Wei-Sheng and Ma, Chao and Zhang, Xiaoyun and Gao, Zhiyong and Yang, Ming-Hsuan},
  booktitle={Proceedings of the IEEE/CVF conference on computer vision and pattern recognition},
  pages={3703--3712},
  year={2019}
}

@inproceedings{jiang2018super,
  title={Super slomo: High quality estimation of multiple intermediate frames for video interpolation},
  author={Jiang, Huaizu and Sun, Deqing and Jampani, Varun and Yang, Ming-Hsuan and Learned-Miller, Erik and Kautz, Jan},
  booktitle={Proceedings of the IEEE conference on computer vision and pattern recognition},
  pages={9000--9008},
  year={2018}
}

@inproceedings{liu2017video,
  title={Video frame synthesis using deep voxel flow},
  author={Liu, Ziwei and Yeh, Raymond A and Tang, Xiaoou and Liu, Yiming and Agarwala, Aseem},
  booktitle={Proceedings of the IEEE international conference on computer vision},
  pages={4463--4471},
  year={2017}
}

@inproceedings{niklaus2018context,
  title={Context-aware synthesis for video frame interpolation},
  author={Niklaus, Simon and Liu, Feng},
  booktitle={Proceedings of the IEEE conference on computer vision and pattern recognition},
  pages={1701--1710},
  year={2018}
}

@inproceedings{hu2022many,
  title={Many-to-many splatting for efficient video frame interpolation},
  author={Hu, Ping and Niklaus, Simon and Sclaroff, Stan and Saenko, Kate},
  booktitle={Proceedings of the IEEE/CVF Conference on Computer Vision and Pattern Recognition},
  pages={3553--3562},
  year={2022}
}

@inproceedings{niklaus2020softmax,
  title={Softmax splatting for video frame interpolation},
  author={Niklaus, Simon and Liu, Feng},
  booktitle={Proceedings of the IEEE/CVF Conference on Computer Vision and Pattern Recognition},
  pages={5437--5446},
  year={2020}
}

@inproceedings{park2020bmbc,
  title={Bmbc: Bilateral motion estimation with bilateral cost volume for video interpolation},
  author={Park, Junheum and Ko, Keunsoo and Lee, Chul and Kim, Chang-Su},
  booktitle={Computer Vision--ECCV 2020: 16th European Conference, Glasgow, UK, August 23--28, 2020, Proceedings, Part XIV 16},
  pages={109--125},
  year={2020},
  organization={Springer}
}

@inproceedings{park2021asymmetric,
  title={Asymmetric bilateral motion estimation for video frame interpolation},
  author={Park, Junheum and Lee, Chul and Kim, Chang-Su},
  booktitle={Proceedings of the IEEE/CVF International Conference on Computer Vision},
  pages={14539--14548},
  year={2021}
}

@inproceedings{sim2021xvfi,
  title={Xvfi: extreme video frame interpolation},
  author={Sim, Hyeonjun and Oh, Jihyong and Kim, Munchurl},
  booktitle={Proceedings of the IEEE/CVF international conference on computer vision},
  pages={14489--14498},
  year={2021}
}

@inproceedings{kong2022ifrnet,
  title={Ifrnet: Intermediate feature refine network for efficient frame interpolation},
  author={Kong, Lingtong and Jiang, Boyuan and Luo, Donghao and Chu, Wenqing and Huang, Xiaoming and Tai, Ying and Wang, Chengjie and Yang, Jie},
  booktitle={Proceedings of the IEEE/CVF Conference on Computer Vision and Pattern Recognition},
  pages={1969--1978},
  year={2022}
}

@inproceedings{lu2022video,
  title={Video frame interpolation with transformer},
  author={Lu, Liying and Wu, Ruizheng and Lin, Huaijia and Lu, Jiangbo and Jia, Jiaya},
  booktitle={Proceedings of the IEEE/CVF Conference on Computer Vision and Pattern Recognition},
  pages={3532--3542},
  year={2022}
}

@inproceedings{shi2022video,
  title={Video frame interpolation transformer},
  author={Shi, Zhihao and Xu, Xiangyu and Liu, Xiaohong and Chen, Jun and Yang, Ming-Hsuan},
  booktitle={Proceedings of the IEEE/CVF Conference on Computer Vision and Pattern Recognition},
  pages={17482--17491},
  year={2022}
}

@misc{long2016learning,
      title={Learning Image Matching by Simply Watching Video}, 
      author={Gucan Long and Laurent Kneip and Jose M. Alvarez and Hongdong Li},
      year={2016},
      eprint={1603.06041},
      archivePrefix={arXiv},
      primaryClass={cs.CV}
}

@inproceedings{flynn2016deepstereo,
  title={Deepstereo: Learning to predict new views from the world's imagery},
  author={Flynn, John and Neulander, Ivan and Philbin, James and Snavely, Noah},
  booktitle={Proceedings of the IEEE conference on computer vision and pattern recognition},
  pages={5515--5524},
  year={2016}
}

@article{kalantari2016learning,
  title={Learning-based view synthesis for light field cameras},
  author={Kalantari, Nima Khademi and Wang, Ting-Chun and Ramamoorthi, Ravi},
  journal={ACM Transactions on Graphics (TOG)},
  volume={35},
  number={6},
  pages={1--10},
  year={2016},
  publisher={ACM New York, NY, USA}
}

@inproceedings{zhou2016view,
  title={View synthesis by appearance flow},
  author={Zhou, Tinghui and Tulsiani, Shubham and Sun, Weilun and Malik, Jitendra and Efros, Alexei A},
  booktitle={Computer Vision--ECCV 2016: 14th European Conference, Amsterdam, The Netherlands, October 11--14, 2016, Proceedings, Part IV 14},
  pages={286--301},
  year={2016},
  organization={Springer}
}

@article{ho2020denoising,
  title={Denoising diffusion probabilistic models},
  author={Ho, Jonathan and Jain, Ajay and Abbeel, Pieter},
  journal={Advances in neural information processing systems},
  volume={33},
  pages={6840--6851},
  year={2020}
}

@inproceedings{ma2025tinyvim,
  title={Tinyvim: Frequency decoupling for tiny hybrid vision mamba},
  author={Ma, Xiaowen and Ni, Zhenliang and Chen, Xinghao},
  booktitle={Proceedings of the IEEE/CVF International Conference on Computer Vision},
  pages={23519--23529},
  year={2025}
}

@article{ibrahim2025survey,
  title={A survey on mamba architecture for vision applications},
  author={Ibrahim, Fady and Liu, Guangjun and Wang, Guanghui},
  journal={arXiv preprint arXiv:2502.07161},
  year={2025}
}

@inproceedings{lyu2025tlb,
  title={TLB-VFI: Temporal-Aware Latent Brownian Bridge Diffusion for Video Frame Interpolation},
  author={Lyu, Zonglin and Chen, Chen},
  booktitle={Proceedings of the IEEE/CVF International Conference on Computer Vision},
  pages={16260--16269},
  year={2025}
}

@inproceedings{lyu2024frame,
  title={Frame interpolation with consecutive brownian bridge diffusion},
  author={Lyu, Zonglin and Li, Ming and Jiao, Jianbo and Chen, Chen},
  booktitle={Proceedings of the 32nd ACM International Conference on Multimedia},
  pages={3449--3458},
  year={2024}
}

@inproceedings{li2023bbdm,
  title={Bbdm: Image-to-image translation with brownian bridge diffusion models},
  author={Li, Bo and Xue, Kaitao and Liu, Bin and Lai, Yu-Kun},
  booktitle={Proceedings of the IEEE/CVF conference on computer vision and pattern Recognition},
  pages={1952--1961},
  year={2023}
}

@inproceedings{sun2018pwc,
  title={Pwc-net: Cnns for optical flow using pyramid, warping, and cost volume},
  author={Sun, Deqing and Yang, Xiaodong and Liu, Ming-Yu and Kautz, Jan},
  booktitle={Proceedings of the IEEE conference on computer vision and pattern recognition},
  pages={8934--8943},
  year={2018}
}

@inproceedings{hui2018liteflownet,
  title={Liteflownet: A lightweight convolutional neural network for optical flow estimation},
  author={Hui, Tak-Wai and Tang, Xiaoou and Loy, Chen Change},
  booktitle={Proceedings of the IEEE conference on computer vision and pattern recognition},
  pages={8981--8989},
  year={2018}
}

@article{montgomery1994xiph,
  title={Xiph. org video test media (derf’s collection)},
  author={Montgomery, Christopher and Lars, H},
  journal={Online, https://media. xiph. org/video/derf},
  volume={6},
  pages={1},
  year={1994}
}

@inproceedings{li2023amt,
  title={Amt: All-pairs multi-field transforms for efficient frame interpolation},
  author={Li, Zhen and Zhu, Zuo-Liang and Han, Ling-Hao and Hou, Qibin and Guo, Chun-Le and Cheng, Ming-Ming},
  booktitle={Proceedings of the IEEE/CVF Conference on Computer Vision and Pattern Recognition},
  pages={9801--9810},
  year={2023}
}
\end{document}